# Learning to Love Edge Cases in Formative Math Assessment:

Using the AMMORE Dataset and Chain-of-Thought Prompting to Improve Grading Accuracy

Owen Henkel, Hannah Horne-Robinson, Maria Dyshel, Nabil Ch, Baptiste Moreau-Pernet, Ralph Abood

**ABSTRACT**

This paper introduces AMMORE, a new dataset of 53,000 math open-response question-answer pairs from Rori, a learning platform used by students in several African countries and conducts two experiments to evaluate the use of large language models (LLM) for grading particularly challenging student answers. The AMMORE dataset enables various potential analyses and provides an important resource for researching student math acquisition in understudied, real-world, educational contexts. In experiment 1 we use a variety of LLM-driven approaches, including zero-shot, few-shot, and chain-of-thought prompting, to grade the 1% of student answers that a rule-based classifier fails to grade accurately. We find that the best-performing approach – chain-of-thought prompting – accurately scored 92% of these edge cases, effectively boosting the overall accuracy of the grading from 98.7% to 99.9%. In experiment 2, we aim to better understand the consequential validity of the improved grading accuracy, by passing grades generated by the best-performing LLM-based approach to a Bayesian Knowledge Tracing (BKT) model, which estimated student mastery of specific lessons. We find that relatively modest improvements in model accuracy at the individual question level can lead to significant changes in the estimation of student mastery. Where the rules-based classifier currently used to grade student, answers misclassified the mastery status of 6.9% of students across their completed lessons, using the LLM chain-of-thought approach this misclassification rate was reduced to 2.6% of students. Taken together, these findings suggest that LLMs could be a valuable tool for grading open-response questions in K-12 mathematics education, potentially enabling encouraging wider adoption of open-ended questions in formative assessment.

## 1. Introduction

Formative assessment and feedback are crucial components of the learning process, enabling students and educators to adapt their approach within or in-between lessons to maximize learning [34]. It has been shown to lead to significant improvements in learning outcomes [18]. Closed-response questions, such as multiple-choice and true/false, are commonly used in formative assessment, and have the benefit of being efficient to grade and can provide instant feedback [31]. However, they have several drawbacks, such as the possibility of students relying on test-taking strategies, a potential lack of face validity, and the complexity of generating multiple answer options [16, 24]. In contrast, open-ended and short answer questions require students to answer a question using their own words often with a few sentences [31]. Many researchers argue that open-response questions decrease the influence of test-taking strategies, have greater face validity, and may be better suited to evaluate certain subprocesses of the skill being assessed [3, 4, 8, 16, 34]. However, the process of grading open-ended questions can be resource-intensive and expensive, which limits their widespread use [23]. While educators may prefer the type of information they can glean from student responses to open-ended questions, the laborious grading process can overburden educators and compromise the quality of feedback, which may limit students' comprehension and critical engagement with the subject matter [25]. Therefore, automatic short answer grading (ASAG) offers a promising solution to address this, but it has historically been challenging to perform easily and effectively enough for widespread use in educational settings [2, 7, 13]. Most state-of-the-art approaches have relied primarily on handcrafted approaches, or more recently fine-tuning models for specific tasks [5, 14], which required extensive technical expertise and large datasets [26, 32].

The field of ASAG has seen significant advancements with the emergence of LLMs, presenting new opportunities for enhancing educational assessment and personalized learning. There is growing evidence that these models can complete evaluation tasks on novel datasets with only minimal prompt engineering [15, 18, 19]. If LLMs can accurately mark open-ended questions, the time savings for educators would be substantial, and could facilitate more frequent and effective formative assessment. However, little is known about how LLMs perform across a variety of educational settings and whether LLMs can be relied upon to generalize to ever more complex use cases. Additionally, there are a limited number of publicly available datasets from educational settings upon which LLMs can be tested. This paper makes two contributions in response to these gaps.

First, we introduce a novel dataset, the African Middle-School Math Open REsponse (AMMORE) dataset, which consists of 53,000 answers to middle school math questions from students in West Africa. The data for AMMORE was collected from Rori, an AI-powered WhatsApp math-tutor that allows students in West Africa to independently practice math concepts free of charge. The dataset's rich structure, which includes question level data, user IDs, learning standard designators and students self-reported age, enables various potential analyses, such as investigating students' skill mastery across micro-lessons, analyzing the relative difficulty of specific questions or micro-lessons across students, or exploring how different grading models' judgments compare to those of humans. This dataset provides a unique opportunity to explore the challenges of grading diverse student responses in a real-world educational context, particularly in regions where access to quality education is often limited and where innovative solutions like AI tutors are being leveraged to bridge educational gaps.

Second, we conduct an extensive empirical evaluation of LLM-based approaches to grade a challenging subset of AMMORE, using a variety of automated approaches. We explore various methods, including string matching, text processing, and different LLM prompting techniques, to evaluate their accuracy and consistency in assessing student responses. We find that LLM-based approaches, particularly chain-of-thought prompting (CoT), outperform traditional methods in grading accuracy, demonstrating their ability to handle the complexity and variability of student responses in open-ended math questions. The superior performance of LLM-based methods is especially evident in cases where students provide correct answers in unexpected formats or use equivalent mathematical expressions. We also explore whether improvement in question grading leads to more accurate estimates of student concept mastery. We find that relatively modest improvements in model accuracy at the individual question level can lead to significant changes in the estimation of student mastery and perceived lesson difficulty. These results have important implications for the design of intelligent tutoring systems (ITS), potentially enabling more accurate adaptive learning pathways and personalized feedback. Our findings also suggest that the use of LLM-based grading could encourage wider adoption of open-ended questions in formative assessment, leveraging their pedagogical benefits without increasing the grading burden on educators.

## 2. Prior Work

**2.1 Automatic Short Answer Grading**

Automatic short answer grading has been an active area of research for over a decade. Burrows et. al. [7] provide a comprehensive overview of approaches up until 2015; while Haller et. al., [17] discuss how ASAG has more recently moved from models based on handcrafted features to approaches including word-embedding and representation learning. However, regardless of the paradigm, most models used for ASAG are explicitly trained or fine-tuned for specific grading tasks [21]. There has been considerable progress with these types of tasks, for instance, Sultan et al. [36] developed a model that represents each sentence as the sum of the individual word embeddings. At the time, this model achieved state-of-the-art performance on the SemEval benchmarking dataset. As these types of models depend on prompt-specific training data, they often need to be re-trained for each individual short answer prompt, which is costly, time-consuming, and in most cases simply infeasible.

The recent rise of ever-larger pre-trained LLMs, trained on vast text corpi, has enabled a new approach: fine-tuning, often referred to as transfer learning. This paradigm of machine learning typically consists of two steps: pre-training and fine-tuning. In the former, a neural network model learns weights through unsupervised learning on a large general dataset. In the latter, the model trains on a smaller, task-specific dataset [20] to update its weights to better align with downstream tasks. As a result of their significant pre-training, LLMs can achieve far better sample efficiency on the target task(s) [6]. For example, Sung et al. [37] fine-tuned BERT, a widely used pre-trained transformer-based language model to grade short-answer responses and found it was able to classify almost at the human-level agreement and achieve superior results to the previous state-of-the-art on the SemEval dataset. More recently, Fernandez et al. [14] used a BERT-based model to evaluate open-response reading comprehension questions and achieved an agreement score with expert raters, as measured by Cohen's Kappa, of 0.84, where human-to-human scores were 0.88.

While pre-trained language models that have been fine-tuned with small task-specific datasets have improved ASAG, their practical application to formative assessment in educational settings remains limited. This is largely due to a few central constraints of this approach: the technical complexity of the fine-tuning process, the continued (albeit small) need for task-specific data, and these models' difficulty in generalizing. First, fine-tuning an LLM requires substantial computational power, which is not widely available in educational contexts. Second, data from educational settings, even in smaller amounts, is remarkably hard to obtain given sensitivities over data sharing and privacy. Finally, LLMs' performance across different tasks is variable and often does not generalize well to different settings, leading to reliability concerns for the overall approach.

**2.2 Potential of Generative LLMs for ASAG**

The current generation of LLMs, including ChatGPT, GPT-4, Claude, Llama, Mistral, Gemini, were trained similarly to previous generations but with significantly larger datasets and a higher number of parameters, in some cases by more than an order of magnitude [9, 35]. Additionally, these models underwent various "instruction fine-tuning" steps to enhance their usability and ability to generalize to new tasks, often with minimal exposure to examples [30]. This also improved their ability to interpret human-written natural language instructions (i.e., prompting), allowing non-technical users to make requests and adapt a model to new tasks by modifying their prompts, rather than requiring further training or fine-tuning [35]. Therefore, it is unsurprising that evidence is growing that LLMs can be used for certain types of grading tasks [21]. Current LLMs can perform various linguistic tasks that previously required the use of task-specific, fine-tuned LLMs [20, 38], and with minimal prompt engineering can complete evaluation tasks on novel datasets [15, 22]. Instead of using a task-specific dataset to fine-tune a pre-trained LLM, a user can now simply write an explanation and a few examples of how they wish the model to grade student answers and achieve reasonable performance.

While there is a growing amount of research on grading essays using generative LLMs, relatively little is known about their potential for ASAG [21, 27, 33]. Morjaria et al. [28] found that ChatGPT graded 6 short answer assessments from an undergraduate medical program similarly to a single expert rater. Cohn et al. [11] found that GPT-4 successfully graded student answers to high school science questions. However, Kortemeyer [21] found that LLMs fell short in certain aspects of grading introductory physics assignments. A review by Schneider et al. [33] concluded that "while 'out-of-the-box' LLMs provide a valuable tool to offer a complementary perspective, their readiness for independent automated grading remains a work in progress."

In all aforementioned cases, the studies were conducted with a small sample of student responses, with a primary focus on high school and university students. However, there has been little exploration of generative LLMs' ability to grade short-answer responses from elementary or middle school students which can be attributed, in part, to the limited number of publicly available short-answer datasets. As a result, we propose our dataset and empirical analysis to help fill this gap in the literature.

## 2.3 Overview of Existing Short Answer Datasets

While there are several math question datasets in the literature (see Table 1 below for a more detailed overview), they present many limitations that undermine their relevance in real-world grading contexts, especially for elementary and middle school students. First, several prominent datasets, e.g., MATH, contain questions and correct answers but do not contain information about how students answered, others, e.g., EEDI, MathE, contain information about students' multiple-choice responses. Second, of the below datasets, only ASSISTments contains information allowing researchers to track progression through a curriculum. Third, few of these datasets contain information from lower resource and underrepresented populations. These limitations are the main motivation behind our proposed dataset AMMORE, which we discuss in more detail in Section 3.

**Table 1**

*Summary of Publicly Available Math Datasets*

| Dataset | Topic | Student Answers / Age | Country | Response Type | Number of Responses |
|---|---|---|---|---|---|
| **MATH** | Competition Mathematics | No / N/A | N/A | Open Response | 12,500 |
| **GSM8K** | Primary School Mathematics | No / N/A | N/A | Open Response | 8,000 + |
| **MathE** | College Mathematics | Yes / No | Multiple | Multiple Choice | 9,546 |
| **COMAT** | Primary & Highschool Mathematics | Yes / No | United States | Conversations & Open Response | 188 |
| **EEDI** | Primary & Highschool Mathematics | Yes / No | United Kingdom | Multiple Choice | 17 million + |
| **NAEP** | Grade 4, Grade 8 Mathematics | Yes / Yes | United States | Constructed Response | 250,000 + |
| **ASSISTMents** | Primary & Highschool Mathematics | Yes/ No | United States | Multiple Choice & Open Response | 1,000,000 + |

# 3. AMMORE Dataset

In this section, we present the African Middle-School Math Open REsponse (AMMORE) Dataset, which contains 53,298 student answers to open response practice questions, assembled from a subset math practice sessions on Rori of 2,508 at-home users that took place between January 1st and April 30th, 2024.

## 3.1 Background

Rising Academies, an educational network based in Ghana, has created Rori, an AI-powered math tutor available on WhatsApp. Rori can be used at home or in schools free of charge. The Rori curriculum has one or more micro-lessons for each skill in the math Global Proficiency Framework (GPF), with over 500 micro-lessons to date. Each micro-lesson includes a brief student-friendly explanation of the skill and ten scaffolded practice questions. Many of these questions require open-ended responses, which was a decision taken for pedagogical reasons. Students are expected to write their answers into WhatsApp using the mobile keyboard. If students answer a question incorrectly, they are first shown a hint to help them solve the question and if their second attempt is unsuccessful, they are shown a worked solution. When students finish a micro-lesson, they are encouraged to continue with the next, which incrementally increases in difficulty. Rori will suggest students move either backwards or forwards in the curriculum if they find a lesson too difficult or easy. For more context you can **watch this 2-minute video**.

      Rori's curriculum is built upon the comprehensive and evidence-based GPF. The framework was developed to create uniform global standards for reading and mathematics across the world and was created by USAID by using inputs from experts representing organizations such as the World Bank, the Bill and Melinda Gates Foundation, the UK's Foreign, Commonwealth, and Development Office, the UNESCO Institute for Statistics, and many more. The GPF represents a global standard for the competencies required for learners at different stages. It covers grades 1 to 9, aligns with national standards globally, and the standards are linked across grade levels. The math framework has five domains: "Numbers and operations", "Measurement", "Geometry", "Statistics and probability", and "Algebra". Each domain is split into constructs, then subconstructs, and then in specific skills that a student in each grade should be able to demonstrate. For example, the domain "Numbers and

operations" has a topic "Integers and Exponents" that has skills such as "Add and subtract" and "Multiply and divide". For a more detailed description of the structure of the curriculum see here.

## 3.2 Structure

Each response in our dataset was scored by a pre-existing, rules-based classification model, native to Rori, which classifies answer attempts as "correct", "wrong" or "other". The latter was typically returned when a student entered something besides an answer attempt, such as a voice note or a sticker. These classifications were then manually reviewed by humans, and changed where necessary, meaning the dataset also has a ground truth score for each student answer. The dataset is comprised of students' answers to math questions from Rori lessons from grade levels 6 to 9 in the domains "Algebra" and "Number and operations". Each student answer is paired with the corresponding question, the expected response, a ground-truth correct/incorrect score, the specific learning standard evaluated by the question, the time the student answered, and a UID number that can be used to link student responses across the dataset.

| Summary Information | | Example attributes of single entry | |
|---|---|---|---|
| Total Answers | 53,031 | lesson | G9.N5.2.1.1 |
| Correct Answers | 34,668 | question_number | *2* |
| Incorrect Answers | 15,278 | question_text | *3^2 + 3^1 = __* |
| Other Answers | 3,085 | expected_answer | 12 |
| Unique Students | 2,508 | student_response | =6+6 =12 |
| Grade Levels Covered | 6-9 | model_grade | wrong |
| Domains Covered | Algebra, Numbers and Operations | human_grade | correct |
| Number of Lessons | 151 | time | 1/9/24 7:57 |
| Number of Skills | 35 | user_id | 17 |

**Figure 1** *Structure of dataset*

The dataset also includes matched but anonymized demographic data on the 2,508 users, such as when they first started using Rori, their country code, self-reported age, and number of messages they sent and active days on Rori. At-home users tend to come from Nigeria, Ghana and South Africa and are mostly between the ages of 10 and 30 and could be using their own or their family members' phones. You can access the AMMORE dataset and data dictionary [here](#).

**3.3 Potential Uses of the AMMORE Rising Dataset**

The dataset's structure enables various potential analyses. For example (a) investigating students' skill mastery across micro-lessons, (b) analyzing the relative difficulty of specific questions or micro-lessons across students, or (c) exploring how the classification model's judgments compare to those of human raters.

Expanding on the first example, while there are many ways to evaluate student mastery at the micro-lesson level, for simplicity, we define mastery as an 80% correct answer rate for questions from a micro-lesson. As discussed above, a micro-lesson is a set of 10 questions of the same difficulty level focusing on a specific learning standard. We consider the responses labelled "correct" or "wrong" and discard those labelled "other" to compute the percentage of micro-lessons that students "mastered". Using this threshold, we can determine that students "mastered" 48% of micro-lessons. To further this analysis, one could combine or "roll up" micro-lesson mastery into skill-level mastery. The dataset includes 151 different micro-lessons covering 35 different skills. For instance, if we posit that a student must master at least 75% of the micro-lessons contained within a skill to have mastered that skill, we can determine how many of the 2,508 students in the dataset have mastered each skill. With this example, 1,133 of the 2,508 students in the data set (45%) would have mastered a skill.

Also, because the same student practices skills at different grade levels, it is possible to compare student age to the grade-level of the topics they are practicing. Using the same mastery thresholds as above, we can determine that amongst the 11% of students who master at least two skills (273 students), 28% of them (76 students) master skills at multiple grade levels. One can also estimate whether students are performing at "grade-level". Our dataset's lessons span grades 6 to 9, with 38% of all answers at level 9, 29% at level 6, then 20% and 13% at levels 7 and 8 respectively.

Yet another approach could be to use this dataset to test different analytics approaches, such as Bayesian Knowledge Tracing (BKT), which we explore in experiment 2, or other mastery prediction models. The rich data available, including question-level responses and progression through micro-lessons over time, makes this dataset particularly suitable for such analyses. These are just a few potential uses for this novel dataset. The combination of detailed student responses, demographic information, and curriculum structure provides a unique opportunity for researchers to explore various aspects of learning analytics, from individual student progress to broader trends in mathematical skill development across grade levels.

## 4. Experiment 1: LLM-based Approaches to Math ASAG

Using a carefully curated subset of challenging student responses from the AMMORE Dataset, we investigate six different automatic grading strategies, ranging from simple string matching to sophisticated LLM-based methods, evaluating their respective performance relative to human scores. We also consider how consistent the models are between repeated runs, if the prompting strategy affects the intra-rater reliability between the model's responses, and how prompting strategy impacts the model response time. Our analysis aims to shed light on the potential of these approaches to improve grading accuracy, particularly when dealing with diverse answer types and formatting variations.

### 4.1 Challenges of Grading Open-Response Math Questions

Accurately grading student answers becomes a complex challenge when moving beyond direct string matches because practice questions on Rori have a diverse set of expected answer types, including fractions, floating-point numbers, and expressions with exponents. In Table 2 you can see a subset of student responses for a given question.

**Table 2**
*Example Student Answers and Labels*

**question_id:** G7.N2.2.3.6

**question_text:** Fill in the missing number: $1/5 \times 2/3 = \_ /15$

**expected_answer:** 2

| student_id | student_answer | model_grade | human_grade |
|---|---|---|---|
| 514 | 2 | correct | correct |
| 1073 | Hold am solving it | other | wrong |
| 876 | is 2 | correct | correct |
| *1203* | *30* | *wrong* | *wrong* |
| *549* | *30/15* | *wrong* | *wrong* |
| *324* | *2/15* | *wrong* | *correct* |

A particular challenge is identifying responses that are correct but have some variation in their formatting or expression that differs from the expected answer. For example, requiring too strict of an answer match would mean only student 514's answer would be accepted, too permissive and only students 1073 and 1203 would be marked incorrect. Accordingly, Rori was already using a relatively sophisticated bespoke classification model to interpret and score student responses, which had already undergone several rounds of improvement. It was this model that generated the initial classifications of student responses for the dataset. However, after human review we found that approximately 1% (1186) of classifications were false negatives, an example of which is student 324 in Table 2.

From a pedagogical perspective, it is important to avoid misclassifying correct student answers (i.e. a false negative) as much as possible - particularly in independent learning environments - as telling a student they made a mistake when they were in fact correct can lead to confusion and frustration. However, false negatives are particularly challenging to identify. For example, looking at the response given by student 324, an expert human reviewer can understand that the student did the core methodical operation correctly and gave the full answer rather than only giving the missing number. While there might be a pedagogical reason to encourage the student to use the correct formatting, treating their answer as wrong would be suboptimal. This contrasts with student 549, who also used the wrong formatting, but performed the operation incorrectly, most likely adding the numerators while multiplying the denominators. Because they require a more sophisticated degree of interpretation, rule-based approaches are unlikely to be successful with these types of student answers. Therefore, we explore the incremental benefits of increasingly sophisticated approaches, combining rule-based systems with LLMs to evaluate the long tail of student answers.

## 4.2 Experimental Design

From the larger AMMORE dataset, we create a smaller dataset, which we refer to as AMMORE-hard. This dataset is comprised of difficult-to-grade student answers, which we used to evaluate the performance of different automatic grading strategies. The resulting dataset comprises of 4,463 answers, including 1528 unique non-trivially correct answers and 2935 unique, non-trivially wrong answers.

AMMORE-hard was created using the following steps: (1) remove answers that were labeled as "other" by a human labeler; (2) remove duplicate occurrences where question, expected answer, and student answer were identical, leaving only one occurrence of each unique combination; (3) remove trivially correct answers, where the student's answer was identical to the expected answer; (4) remove trivially wrong answers, where the expected answer was one character long and the student's answer was one character long (mostly multiple-choice questions with wrong answer); (5) remove trivially wrong answers, where the student's answer was an integer different from the integer expected; and finally, (6) remove answers where either the question was ambiguous or the expected answer was wrong. Using AMMORE-hard, six approaches were used to classify a student answer as correct or wrong.

| | |
|---|---|
| **"Naive" string matching** | Simple rule-based evaluation of matching the expected answer with the student's response. |
| **Text processing** | Evaluation with additional text substitutions and symbolic evaluations. |
| **LLM Zero-shot prompting** | Evaluation using an LLM prompt without specific examples. |
| **LLM Few-shot prompting** | Evaluation using an LLM prompt with a small set of examples. |
| **LLM Chain-of-thought prompting** | Evaluation using an LLM prompt instructing it to show its reasoning process. |
| **Naive string matching, text processing, and zero-shot prompting** | Evaluation proceeded through the evaluations until a correct answer was found or all three evaluations had run: simple rule-based evaluation, text substitutions and symbolic evaluations, and an LLM prompt without examples. |

**Figure 2** *Different approaches to grading student answers*

To make a prediction, each approach was given the same information from the dataset: the question text, the expected answer to the question, and the student's response. The evaluation approach would predict if the answer is "correct" or "wrong". The resulting prediction was recorded. At the time of writing, the model with the strongest performance score on math benchmarks is OpenAI's GPT-4o. Hence, each experiment of a prompt approach used GPT-4o as the LLM. Its temperature setting was set to 0 to reduce the variability of model outputs. No student demographic information was fed to the LLM, nor was it shown the human labels of a student answer.

## 4.2.1 Prompting Strategy

We employ a relatively simple prompting strategy, as the task is straightforward. The base part of the prompt was similar across all strategies. The zero-shot prompt included a description of the core task and slots for the dataset values. The few-shot prompt added three examples of correct answers. These examples represented common student response patterns of equivalent answers: 1) where a student wrote the answer and 2) where a student wrote out their work to arrive at the answer. Instead of providing examples, the chain-of-thought prompt instructed the model to think step-by-step and present a rationale for the classification chosen. The chain-of-thought evaluation used the DSPy framework, which dynamically created a chain-of-thought prompt. Figure 3 shows the prompts for each strategy.

| **Zero-shot Prompt** | **Few-shot Prompt** | **Chain-of-thought Prompt** |
|---|---|---|
| You are a math assistant. You are evaluating whether a student's submission to a math question is right or wrong. The student may have submitted a correct answer in a variety of acceptable, equivalent ways. You must tell whether their submission correctly solves the problem or whether their submission contains a valid answer that is equivalent to the expected answer. If the student's submission is correct or equivalent, write "yes". If the submission is incorrect and not equivalent, write "no". You should only write "yes" or "no".<br><br>## Question<br>{question}<br><br>## Expected Answer<br>{expected_answer}<br><br>## Student Submission<br>{student_message} | You are a math assistant. You are evaluating whether a student's submission to a math question is right or wrong. The student may have submitted a correct answer in a variety of acceptable, equivalent ways. You must tell whether their submission correctly solves the problem or whether their submission contains a valid answer that is equivalent to the expected answer. If the student's submission is correct or equivalent, write "yes". If the submission is incorrect and not equivalent, write "no". You should only write "yes" or "no".<br><br>## Examples<br>### Example 1 - The student gave their work and showed the correct answer.<br>- Question: Solve for z in the proportion: 9/3 = 27/z.<br>- Expected Answer: 9<br>- Student Submission: 9/3=27/a.9×z=3×27.9z/9=91/9.z=9<br>- is_correct: yes<br><br>### Example 2 - The student wrote the correct answer option and its value.<br>- Question: 9 / ___ = 0.25 A) 18 B) 36 C) 81 D) 72<br>- Expected Answer: B<br>- Student Submission: B.36<br>- is_correct: yes<br><br>## Question<br>{question}<br><br>## Expected Answer<br>{expected_answer}<br><br>## Student Submission<br>{student_message} | You are a math assistant. You are evaluating whether a student's submission to a math question is right or wrong. The student may have submitted a correct answer in a variety of acceptable, equivalent ways. You must tell whether their submission correctly solves the problem or whether their submission contains a valid answer that is equivalent to the expected answer.<br><br>Follow the following format.<br><br>Question: the math question<br>Expected Answer: the student's response to the question<br>Reasoning: Let's think step by step in order to produce the correct answer<br><br>We...<br><br>Answer: correct_answer if the student correctly solves the problem or whether their submission contains a valid answer that is equivalent to the expected answer, wrong_answer otherwise<br><br>Question: {question}<br>Expected Answer: {expected_answer}<br>Student Answer: {student_answer}<br>Reasoning: Let's think step by step in order to solve the equation {question} |

**Figure 3** *System Prompts Used in Experiment*

To establish a baseline and evaluate the individual prompt strategies, we created a script that called the relevant functions from Rori's answer evaluation API. The script pulled the question, expected answer,

F. Author et al.

and student answer from the dataset. For the baseline evaluation, the script only ran these values through various string processing strategies. For each prompt strategy evaluation, the script inserted the values into appropriate parts of the prompt and passed the complete prompt to the OpenAI API. The script recorded all evaluation run responses (i.e. the predicted class). Access to prompts and script can be found [here](#).

## 4.3 Results

Table 3 shows the results of the six approaches. As mentioned earlier, each answer evaluation would label a student's answer as "correct" or "wrong". These predictions were compared against the label assigned by a human rater. In Table 3, a result closer to one indicates that the human label and the prediction were similar (i.e., both labeled a student answer as "wrong_answer"). A lower score would indicate that the human label and the predicted label differed (i.e., the human label marked "correct_answer" and the predicted label "wrong_answer").

We report a set of widely used metrics in classification problems which measure model performance after accounting for imbalanced classes in the dataset: precision, recall, and F1 score (Banerjee et al., 2008). We also report the Kappa scores, which are chance-adjusted metrics of agreement, with values ranging from -1 to 1. A value of 1 indicates perfect agreement, 0 suggests that the agreement is only what would be expected by chance, and a value less than 0 indicates agreement worse than random chance. While there are several different measures of chance-adjusted agreement, because we are evaluating 2-class ratings (wrong/correct), we use Linear Weighted Kappa (LWK).

**Table 3**
*Performance of Answer Evaluation Approaches on 2-Class Task*

|                    | Prediction | Accuracy | Precision | Recall | F1   | LWK  |
|--------------------|------------|----------|-----------|--------|------|------|
| **String Matching** | *Wrong*    | 0.79     | 0.76      | 0.99   | 0.86 | 0.44 |
|                    | *Correct*  | 0.79     | 0.97      | 0.39   | 0.56 |      |
| **Text processing** | *Wrong*    | 0.96     | 0.96      | 0.97   | 0.97 | 0.90 |
|                    | *Correct*  | 0.96     | 0.94      | 0.93   | 0.94 |      |
| **LLM Zero-shot**   | *Wrong*    | 0.94     | 0.93      | 0.98   | 0.95 | 0.86 |
|                    | *Correct*  | 0.94     | 0.96      | 0.85   | 0.90 |      |
| LLM Few-shot | *Wrong* | 0.93 | 0.91 | 0.99 | 0.95 | 0.83 |
| | *Correct* | 0.93 | 0.97 | 0.81 | 0.88 | |
| LLM Chain-of-thought | *Wrong* | 0.97 | 0.97 | 0.98 | 0.98 | 0.93 |
| | *Correct* | 0.97 | 0.96 | 0.94 | 0.95 | |
| Text process + LLM Zero-shot | *Wrong* | 0.96 | 0.97 | 0.97 | 0.97 | 0.92 |
| | *Correct* | 0.96 | 0.94 | 0.95 | 0.95 | |

All evaluation techniques generally performed better when attempting to predict a wrong answer rather than the correct answer. Lower recall and F1 scores in the zero-shot and few-shot results may indicate that these techniques have trouble handling the variety of unique correct answer formats in the dataset. This led to more false negatives as a result, which as mentioned is not ideal for educational scenarios. Surprisingly, enhancing the prompt with examples did not necessarily lead to improved performance. Evaluating a student message using a combination of text processing and LLM zero-shot prompting had similar performance to the chain-of-thought prompting.

The performance of chain-of-thought prompting points to improved capability of latest LLMs to deal with mathematical tasks and solve math problems. Many of the model's verdicts can therefore be attributed to differences in pedagogical standards, such as the evaluation presented in Figure 4.

| | |
|---|---|
| **question_id:** | G6.N1.2.2.1.7 |
| **question_text :** | $3^2 + 3^1 = \_\_$ |
| **expected_answer:** | 12 |
| **student_response:** | =6+6 |
| | =12 |
| **human_category:** | correct_answer |
| **llm_cot_category:** | wrong_answer |

| llm_cot_rationale: | We need to evaluate the student's response to see if it correctly solves the problem or contains a valid answer that is equivalent to the expected answer. |
|---|---|
| | First, let's solve the problem step by step:<br>- Calculate $(3^2)$: $(3^2 = 9)$<br>- Calculate $(3^1)$: $(3^1 = 3)$<br>- Add the results: $(9 + 3 = 12)$ |
| | The expected answer is 12. |
| | Now, let's examine the student's answer:<br>- The student wrote "=6+6 =12" |
| | The student's answer is not a correct step-by-step solution to the problem. The student incorrectly broke down the calculation into $(6 + 6)$, which is not how $(3^2 + 3^1)$ should be computed. |

**Figure 4** *Output of different LLM (GPT-4o) strategies on a sample question-answer pair*

While a more lenient educator might credit the student for arriving at the correct answer, the model chose to rate the response as wrong due to incorrect solution steps. Similarly, the model chose to mark wrong responses that contained Unicode analogs of the expected answer (e.g. À instead of A), badly formatted numbers (e.g. 9.34.5 instead of 9.345), and similar small variations that human labelers tended to classify as "correct", taking into account the student's limitations when using a mobile keyboard to type an answer. Looking further into chain-of-thought, the responses reveals that the model will sometimes make a mistake in its mathematical reasoning. This can lead to the model ignoring the expected answer, resulting in an inaccurate evaluation. Figure 5 provides an example of this behavior.

| question_id : | G6.N1.2.2.1.7 |
|---|---|
| question_text : | What is the value of 7 in 4,170,348?<br>A. 7,000,000<br>B. 7,000<br>C. 700,000<br>D. 70,000 |
| expected_answer: | D |
| student_response: | d |
| human_grade: | correct_answer |
| llm_cot_grade: | wrong_answer |

| **llm_cot_rationale:** | We need to determine the place value of the digit 7 in the number 4,170,348. |
|---|---|
| | 1. The digit 7 is in the hundred thousand place. |
| | 2. The value of a digit in the hundred thousand place is the digit multiplied by 100,000. |
| | 3. Therefore, the value of 7 in 4,170,348 is 7 * 100,000 = 700,000. |
| | The correct answer is C (700,000). The student's answer is d, which corresponds to 70,000. |

**Figure 5** *Example of LLM with CoT prompt containing faulty mathematical reasoning*

### 4.3.1 Performance vs Latency

Table 4 shows the average and longest processing times each evaluation took to make a prediction. While chain-of-thought prompting resulted in small but stable improvements over the string processing and symbolic evaluations, it also significantly increased response latency. On average, chain-of-thought responses took 2.79 seconds, compared to 0.73 seconds for few-shot LLM calls. The few-shot evaluation took slightly longer than the zero-shot approach. Text processing evaluations took considerably less time than all prompt-based approaches, which is expected given that this approach did not require connection to the model over internet or the execution of a large-scale machine learning model.

**Table 4**
*Latency of Four Answer Evaluation Approaches on 2-Class Task in Seconds*

|  | **Average Processing Time** | **Longest Processing Time** |
|---|---|---|
| **Text Processing** | *0.006* | *0.269* |
| **LLM Zero-shot** | *0.68* | *5.687* |
| **LLM Few-shot** | *0.73* | *5.937* |
| **LLM Chain-of-thought** | *2.79* | *16.281* |

These results indicate that LLM processing time can be affected by the amount of input tokens the model needs to consume in the case of a longer prompt (such as in a few-shot prompts), and can be increased significantly when the model needs to generate a significant amount of output tokens (such as in the case of chain-of-thought prompting). Additionally, prompt-based approaches could experience more fluctuation in processing time. String processing and symbolic evaluation can reasonably good performance with less latency and more consistent processing time.

### 4.3.2 Model Reliability

While the deterministic approaches like text processing provide consistent results, generative LLMs generate their output using probabilistic methods, and therefore can return different outputs given the same inputs. This variation may occur even when the temperature is set to 0. In some respects, this is similar to human raters, who occasionally will award different ratings to the same student response, when asked to re-rate it after a period of time. Measures of intra-rater reliability are intended to evaluate the extent to which a single rater agrees with their own judgment over time.

To investigate the consistency of prompt-based methods, zero-shot and chain-of-thought approaches were rerun 10 times on a smaller dataset of 100 examples. As shown earlier, these two approaches scored the highest of the prompt-based approaches. For each run, the model labels were compared against the predicted labels to get a Fleiss's Kappa score to measure inter-rater reliability for the run. Table 5 shows the results of these runs. All runs were then compared against each other to arrive at a Fleiss Kappa to represent inter-run reliability.

**Table 5**
*Agreement Between Model Runs and Human Labeling Using Fleiss's Kappa*

|  | Run 1 | Run 2 | Run 3 | Run 4 | Run 5 | Run 6 | Run 7 | Run 8 | Run 9 | Run 10 | Fleiss's Kappa |
|---|---|---|---|---|---|---|---|---|---|---|---|
| **LLM Zero-shot** | 0.66 | 0.66 | 0.68 | 0.70 | 0.66 | 0.62 | 0.66 | 0.70 | 0.66 | 0.66 | 0.90 |
| **LLM Chain-of-thought** | 0.86 | 0.72 | 0.74 | 0.74 | 0.74 | 0.72 | 0.74 | 0.66 | 0.70 | 0.72 | 0.88 |

Both chain-of-thought and zero-shot approaches had relatively high inter-run reliability as measured by Fleiss Kappa. However, the results indicate that chain-of-thought grading, while showing higher answer validity (represented by higher agreement with human labeler), has lower reliability between individual run outcomes and increased possibility for "outlier" runs, occasionally scoring worse than few-shot prompting.

This suggests that chain-of-thought prompting may experience more variation in how it scores responses, which may stem from its reasoning differing between runs. It also indicates that the LLM's "pedagogical standard" may be less consistent. This could lead to accepting answers with typographical errors or other discrepancies outlined earlier, while rejecting them in other instances. While a student may not answer the same question multiple times, this variation could cause student confusion when the LLM does not consistently handle a particular answer pattern (such as substituting Unicode characters).

## 5. Experiment 2: Impact of Improved Grading on Student Ability Estimates

While improving model performance in grading short answer questions is an important area of research, we also seek to better understand the impact of such models on the analysis of student learning. In our second experiment, we investigate whether improved accuracy in model grading corresponded to changes in our estimates of student ability. In the context of a learning environment, even a small number of misgraded answers can lead to vastly different judgments of student ability when aggregated across questions. Tracking a student's progress and understanding of the subject is an essential part of ITS [1, 10] . Accurately estimating a student's current knowledge state enables these systems to deliver a personalized learning experience. For example, student modeling can be used by ITS for making key decisions such as which problem a student should attempt, how much practice is needed to master a skill before moving to a more advanced topic, and when to provide immediate feedback to struggling students.

Bayesian Knowledge Tracing [12] is one of the most widely used algorithms to model students' knowledge in ITS [1]. At any given moment BKT assumes that when a student attempts to demonstrate a skill, they either know the skill or not. Every time a student attempts to demonstrate the skill, the probability of them knowing the skill is updated based on their performance up to that point and whether they were able to demonstrate the skill correctly or not.

Standard BKT uses four parameters to model student knowledge. Two parameters are related to learners' knowledge. When first attempting to demonstrate a skill, a student has the initial probability $P(L_0)$ of knowing the skill. This probability is updated each time the student attempts to demonstrate the skill (i.e., after t attempts, the probability of knowing the skill is $P(L_t)$). At each practice opportunity, a student has a probability $P(T)$ of learning the skill. The other two BKT parameters are related to learners' performance. The probability of a student knowing the skill and yet making a mistake when attempting to demonstrate the skill is $P(S)$. $P(G)$ represents the probability of a student correctly guessing the answer even when not knowing the skill.

## 5.1 Methodology

To quantify the effect of different automated grading algorithms on predicting individual student mastery, we apply the algorithms described in the previous section to generate answer correctness labels for the entire dataset. We exclude questions labeled by human annotators as "other", as there are no straightforward ways to incorporate student non-attempts into the BKT evaluation.

We calculate BKT scores for each student on every lesson they attempted, using only their first attempts to respond to each question. To calculate these scores, we use the following default parameters for every lesson, as suggested by Nguyen et al., [29] : P(L0)=0.4, P(T)=0.05, P(S)=0.299, and P(G)=0.299. To determine if a student had mastered a lesson, we use the last BKT score calculated for that student in each lesson. While mastery thresholds for BKT scores vary between different sources, we choose a threshold of 0.9 to signify that a student had mastered the lesson.

Next, to investigate the effect of grading mechanisms on evaluating individual student mastery, we calculate the number of lessons each student mastered according to different grading algorithms. We then compare these numbers between the worst-performing algorithm (naive string matching) and the best-performing algorithm (chain-of-thought), using human labels of the students' answers as the gold standard. Finally, to estimate the effect of grading mechanisms on evaluating micro-lesson difficulty, we calculate the median mastery score for each micro-lesson and compared this measure across different grading approaches.

## 5.2 How Grading Methods Impact Mastery Predictions

When comparing the number of lessons that reach our threshold for mastery (BKT score of 0.9) according to different grading approaches, we find that 6.9% (165 out of 2,388) of students had their mastery of a completed lesson incorrectly estimated. In contrast, the most successful grading approach, LLM chain-of-thought grading, only underestimated the number of completed lessons for 2.6% of students (61 out of 2,388) students.

This difference can be illustrated by looking at a specific lesson, G7.N3.2.2.2, which demonstrates how dramatic the effect of grading approach on lesson difficulty estimation can be. This lesson deals with changing forms and asks the student to present a given decimal number as a fraction. As there are multiple correct answers to this question and string-matching evaluation struggles with identifying equivalent fractions, the string-matching algorithm would regularly grade mathematically correct results as wrong. Examples of this difference in terms of a single lesson can be seen in Table 6.

**Table 6**
*Change in BKT Score on Lesson G7.N3.2.2.2 by Grading Method for Example Students*

| user_id | BKT Estimate with String Match Grading | BKT Estimate with LLM CoT Grading | BKT Estimate with Human Grading |
|---|---|---|---|
| 996 | 0.349435 | 0.845858 | 0.845858 |
| 1165 | 0.629638 | 0.966567 | 0.966567 |
| 1235 | 0.173999 | 0.809262 | 0.809262 |
| 1239 | 0.895698 | 0.973051 | 0.973051 |
| 1841 | 0.128321 | 0.913219 | 0.913219 |
| 2037 | 0.295264 | 0.994347 | 0.994347 |

Anecdotally, we observe that while the overall number of misgraded answers by simpler methods like string-matching was relatively small, these errors tended to be concentrated around certain students or specific lessons. Students who adapted more slowly to the expected answer format were disproportionately affected by inaccurate grading. Additionally, certain lessons that allowed for multiple correct answer formats or required understanding of equivalent expressions (such as fractions) seemed to be more susceptible to grading errors from simpler methods. For one student, 1190, using string-matching to grade their answers resulted in BKT estimating that they mastered zero lessons, while both human and LLM-based grading resulted in BKT estimate of over 0.90 for all the lesson they completed.

Another interesting specific case demonstrates the impact of inaccurate grading on both student experience and behavior, as well as mastery estimation. Student 994 began their practice with multiple-choice questions in lesson G6.N1.3.6.1. However, because they were not following the expected answer format, their correct answers were graded as wrong. This presumably caused the student to abandon the lesson midway and start a different lesson, where the situation repeated itself. The student then switched to another lesson again after just 3 questions. However, once they started a lesson where the answer format was less ambiguous, the grading quality improved. From that point on, not only did the student start completing the lessons, solving all 10 questions, but the estimation of their mastery also became more aligned with their actual performance.

## 6. Discussion

### 6.1 Implications

The results of our experiments have significant implications for the field of ASAG and its application in educational settings. The superior performance of LLM-based approaches, particularly chain-of-thought prompting, suggests that these models can effectively handle the complexity and variability of student responses in open-ended math questions.

One of the most important implications of our findings is the potential for more widespread use of open-ended questions in formative assessment. As noted in the introduction, open-ended questions have several advantages over closed-response formats, including decreased influence of test-taking strategies and greater face validity. The ability to accurately grade these questions automatically could encourage educators to incorporate more open-ended questions into their assessments, potentially leading to more effective evaluation of student understanding and improved learning outcomes.

The improved accuracy of LLM-based grading approaches also has implications for student experience and engagement. As demonstrated in our analysis of individual student mastery prediction, inaccurate grading can significantly impact a student's perceived progress and potentially influence their behavior. More accurate grading could lead to better alignment between a student's actual performance and their estimated mastery, potentially increasing motivation and reducing frustration. Furthermore, the ability of LLMs to handle diverse answer formats and equivalent expressions could promote more flexible problem-solving among students. Instead of being constrained to a specific answer format, students could express their solutions in ways that feel most natural to them, knowing that the grading system can accurately evaluate their responses.

The implications extend beyond the design and implementation of ITS as well. From a resource perspective, the ability to accurately grade open-ended questions automatically could lead to significant time savings for educators. This could allow them to focus more on providing personalized feedback and support rather than spending time on routine grading tasks.

## 6.2 Limitations

Despite the promising results, our study has several limitations that should be considered when interpreting the findings and planning future research. Firstly, our dataset is limited to middle school mathematics questions from specific domains ("Algebra" and "Numbers and operations"). The performance of the grading approaches, particularly the LLM-based methods, may vary for different subject areas, complexity levels, or age groups.

Secondly, our experiments focused on a binary classification of answers as correct or incorrect. This simplification, while useful for our analysis, does not capture the full spectrum of partial understanding that students may demonstrate in their responses. A more nuanced grading approach might provide richer insights into student comprehension and learning progress.

Thirdly, LLM-based approaches revealed some inconsistency in grading, particularly for the chain-of-thought method. This variability in "pedagogical standards" between runs could be problematic in educational settings where consistent evaluation is crucial for fair assessment and student trust in the system. Relatedly, there is the potential for LLM hallucination or faulty mathematical reasoning, as demonstrated in some of our examples. While these instances were relatively rare, they highlight the need for caution when relying solely on LLM-based grading without human oversight.

Lastly, while we demonstrate the impact of grading accuracy on estimates of student mastery and lesson difficulty, we did not explore how these improved estimates might translate into better learning outcomes in practice. The real-world educational impact of using LLM-based grading in an ITS remains to be studied.

## 6.3 Further Research

Our findings open several avenues for future research in the field of ASAG and its applications in education. One crucial area for further investigation is the expansion of available datasets to include a wider range of subjects, grade levels, and cultural contexts. This would allow researchers to test the generalizability of LLM-based grading approaches across different educational domains and student populations. Additionally, creating datasets that include more complex, multi-step problems could help push the boundaries of what automatic grading systems can handle. Future studies should also explore more nuanced grading scales beyond binary classification. Developing and evaluating methods for assigning partial credit or identifying specific misconceptions in student responses could provide more detailed insights into student understanding and learning progress.

Research into improving the consistency of LLM-based grading is another important direction. This could involve experimenting with different prompting strategies, exploring ensemble methods that combine multiple LLM runs, or investigating ways to fine-tune LLMs for more consistent performance in educational grading tasks. Another promising direction is the integration of LLM-based grading into ITS and studying its impact on adaptive learning. Researchers could investigate how more accurate grading and mastery estimation influence the effectiveness of personalized learning paths, problem selection, and intervention strategies.

Finally, research into hybrid approaches that combine the strengths of rule-based systems, traditional machine learning, and LLMs could lead to more robust and efficient grading systems. This could involve developing frameworks that can dynamically select the most appropriate grading method based on the specific characteristics of each question and response. By pursuing these research directions, we can continue to advance the field of automatic short answer grading and work towards more effective, fair, and personalized educational technologies that support both students and educators in the learning process.

## 7. Conclusion

We make two contributions to the fields of ASAG and LLM evaluation. By presenting AMMORE, we aim to expand and diversify the range of publicly available datasets. As it includes students from Africa answering math questions at middle school levels and provides demographic data, it is a unique dataset that enables a variety of analyses, a few of which we have explored here. We find that chain of thought prompting is the best LLM-driven approach to grade open-response math answers. Additionally, we find improving grading accuracy can lead to significant changes in the estimation of student mastery, which could have considerable impact on the field of ITS and opens up many more questions for future research.

**Declarations and Acknowledgement**

The first author has an ongoing research partnership with Rising Academies and works as a consultant on a project related to developing a conversational agent to support students' math skills. Author 2 works for Rising Academies as Research and Assessment Manager. The remaining authors all work as research consultants on projects with Rising Academies. We would like to thank John Whitmer and Alexis for their support in assembling the AMMORE dataset. We would also like to thank Ryan Baker for guidance on BKT modeling.

# REFERENCES

[1] Abdelrahman, G., Wang, Q. and Nunes, B. 2023. Knowledge Tracing: A Survey. *ACM Computing Surveys*. 55, 11 (Nov. 2023), 1–37. DOI:https://doi.org/10.1145/3569576.

[2] Allen, L.K., Snow, E.L., Crossley, S.A., Tanner Jackson, G. and McNamara, D.S. 2014. Reading comprehension components and their relation to writing. *L'Année psychologique*. 114, 04 (Dec. 2014), 663–691. DOI:https://doi.org/10.4074/S0003503314004047.

[3] van den Bergh, H. 1990. On the Construct Validity of Multiple- Choice Items for Reading Comprehension. *Applied Psychological Measurement*. 14, 1 (Mar. 1990), 1–12. DOI:https://doi.org/10.1177/014662169001400101.

[4] Black, P. and Wiliam, D. 2009. Developing the theory of formative assessment. *Educational Assessment, Evaluation and Accountability*. 21, 1 (2009), 5–31. DOI:https://doi.org/10.1007/s11092-008-9068-5.

[5] Botelho, A., Baral, S., Erickson, J.A., Benachamardi, P. and Heffernan, N.T. 2023. Leveraging natural language processing to support automated assessment and feedback for student open responses in mathematics. *Journal of Computer Assisted Learning*. 39, 3 (Jun. 2023), 823–840. DOI:https://doi.org/10.1111/jcal.12793.

[6] Brown, T.B. et al. 2020. Language Models are Few-Shot Learners. *arXiv:2005.14165 [cs]*. (Jul. 2020).

[7] Burrows, S., Gurevych, I. and Stein, B. 2015. The Eras and Trends of Automatic Short Answer Grading. *International Journal of Artificial Intelligence in Education*. 25, 1 (2015), 60–117. DOI:https://doi.org/10.1007/s40593-014-0026-8.

[8] Cain, K. and Oakhill, J. 2007. *Children's comprehension problems in oral and written language a cognitive perspective*. Guilford Press.

[9] Chowdhery, A. et al. 2023. PaLM: Scaling Language Modeling with Pathways. (2023).

[10] Chrysafiadi, K. and Virvou, M. 2013. Student modeling approaches: A literature review for the last decade. *Expert Systems with Applications*. 40, 11 (Sep. 2013), 4715–4729. DOI:https://doi.org/10.1016/j.eswa.2013.02.007.

[11] Cohn, C., Hutchins, N., Le, T. and Biswas, G. 2024. A Chain-of-Thought Prompting Approach with LLMs for Evaluating Students' Formative Assessment Responses in Science. arXiv.

[12] Corbett, A.T. and Anderson, J.R. 1995. Knowledge tracing: Modeling the acquisition of procedural knowledge. *User Modelling and User-Adapted Interaction*. 4, 4 (1995), 253–278. DOI:https://doi.org/10.1007/BF01099821.

[13] Crossley, S.A., Kim, M., Allen, L. and McNamara, D. 2019. Automated Summarization Evaluation (ASE) Using Natural Language Processing Tools. *Artificial Intelligence in Education*. S. Isotani, E. Millán, A. Ogan, P. Hastings, B. McLaren, and R. Luckin, eds. Springer International Publishing. 84–95.

[14] Fernandez, N., Ghosh, A., Liu, N., Wang, Z., Choffin, B., Baraniuk, R. and Lan, A. 2023. Automated Scoring for Reading Comprehension via In-context BERT Tuning. arXiv.

[15] Gilardi, F., Alizadeh, M. and Kubli, M. 2023. ChatGPT outperforms crowd workers for text-annotation tasks. *Proceedings of the National Academy of Sciences*. 120, 30 (Jul. 2023), e2305016120. DOI:https://doi.org/10.1073/pnas.2305016120.

[16] Gurung, A., Vanacore, K., Mcreynolds, A.A., Ostrow, K.S., Worden, E., Sales, A.C. and Heffernan, N.T. 2024. Multiple Choice vs. Fill-In Problems: The Trade-off Between Scalability and Learning. *Proceedings of the 14th Learning Analytics and Knowledge Conference* (Kyoto Japan, Mar. 2024), 507–517.

[17] Haller, S., Aldea, A., Seifert, C. and Strisciuglio, N. 2022. Survey on Automated Short Answer Grading with Deep Learning: from Word Embeddings to Transformers. arXiv.

[18] Hattie, J. 2010. *Visible learning: a synthesis of over 800 meta-analyses relating to achievement*. Routledge.

[19] Henkel, O., Hills, L., Roberts, B. and McGrane, J. 2023. Supporting Foundational Literacy Assessment in LMICs: Can LLMs Grade Short-answer Reading Comprehension Questions? (2023).

[20] Kojima, T., Gu, S.S., Reid, M., Matsuo, Y. and Iwasawa, Y. 2022. Large Language Models are Zero-Shot Reasoners. (2022).

[21] Kortemeyer, G. 2023. Performance of the Pre-Trained Large Language Model GPT-4 on Automated Short Answer Grading. arXiv.

[22] Kuzman, T., Mozetič, I. and Ljubešić, N. 2023. ChatGPT: Beginning of an End of Manual Linguistic Data Annotation? Use Case of Automatic Genre Identification. arXiv.
[23] Magliano, J.P. and Graesser, A.C. 2012. Computer-based assessment of student-constructed responses. *Behavior Research Methods*. 44, 3 (2012), 608–621. DOI:https://doi.org/10.3758/s13428-012-0211-3.
[24] Magliano, J.P. and Millis, K.K. 2003. Assessing Reading Skill With a Think-Aloud Procedure and Latent Semantic Analysis. *Cognition and Instruction*. 21, 3 (2003), 251–283. DOI:https://doi.org/10.1207/S1532690XCI2103_02.
[25] Matelsky, J.K., Parodi, F., Liu, T., Lange, R.D. and Kording, K.P. 2023. A large language model-assisted education tool to provide feedback on open-ended responses. arXiv.
[26] Mayfield, E. and Black, A.W. 2020. Should You Fine-Tune BERT for Automated Essay Scoring? *Proceedings of the Fifteenth Workshop on Innovative Use of NLP for Building Educational Applications* (Seattle, WA, USA → Online, 2020), 151–162.
[27] Mizumoto, A. and Eguchi, M. 2023. Exploring the potential of using an AI language model for automated essay scoring. *Research Methods in Applied Linguistics*. 2, 2 (Aug. 2023), 100050. DOI:https://doi.org/10.1016/j.rmal.2023.100050.
[28] Morjaria, L., Burns, L., Bracken, K., Levinson, A.J., Ngo, Q.N., Lee, M. and Sibbald, M. 2024. Examining the Efficacy of ChatGPT in Marking Short-Answer Assessments in an Undergraduate Medical Program. *International Medical Education*. 3, 1 (Jan. 2024), 32–43. DOI:https://doi.org/10.3390/ime3010004.
[29] Nguyen, H.A., Hou, X. and Stamper, J. 2020. Moving beyond Test Scores: Analyzing the Effectiveness of a Digital Learning Game through Learning Analytics. (2020).
[30] Ouyang, L. et al. 2022. Training language models to follow instructions with human feedback. arXiv.
[31] Pearson, P.D. and Hamm, D.N. 2006. The Assessment of Reading Comprehension: A Review of Practices—Past, Present, and Future. *Children's reading comprehension and assessment*. Lawrence Erlbaum Associates.
[32] Pulman, S.G. and Sukkarieh, J.Z. 2005. Automatic short answer marking. *Proceedings of the second workshop on Building Educational Applications Using NLP - EdAppsNLP 05* (Ann Arbor, Michigan, 2005), 9–16.
[33] Schneider, J., Schenk, B., Niklaus, C. and Vlachos, M. 2023. Towards LLM-based Autograding for Short Textual Answers. (2023).
[34] Shute, V.J. 2008. Focus on Formative Feedback. *Review of Educational Research*. 78, 1 (Mar. 2008), 153–189. DOI:https://doi.org/10.3102/0034654307313795.
[35] Stiennon, N., Ouyang, L., Wu, J., Ziegler, D.M., Lowe, R., Voss, C., Radford, A., Amodei, D. and Christiano, P. 2022. Learning to summarize from human feedback. arXiv.
[36] Sultan, M.A., Salazar, C. and Sumner, T. 2016. Fast and Easy Short Answer Grading with High Accuracy. *Proceedings of the 2016 Conference of the North American Chapter of the Association for Computational Linguistics: Human Language Technologies* (San Diego, California, 2016), 1070–1075.
[37] Sung, C., Dhamecha, T., Saha, S., Ma, T., Reddy, V. and Arora, R. 2019. Pre-Training BERT on Domain Resources for Short Answer Grading. *Proceedings of the 2019 Conference on Empirical Methods in Natural Language Processing and the 9th International Joint Conference on Natural Language Processing (EMNLP-IJCNLP)* (Hong Kong, China, 2019), 6070–6074.
[38] Wei, J. et al. 2022. Emergent Abilities of Large Language Models. arXiv.